\documentclass[
]{ceurart}

\usepackage{listings}
\usepackage{booktabs}
\usepackage{amsmath}
\usepackage{graphicx}
\usepackage{caption}
\usepackage{subcaption}
\usepackage{algorithm}
\usepackage{algorithmic}
\usepackage{todonotes}

\usepackage{tabularx}
\usepackage{makecell}

\begin{document}

\copyrightyear{2026}
\copyrightclause{Copyright for this paper by its authors.
  Use permitted under Creative Commons License Attribution 4.0
  International (CC BY 4.0).}

\conference{CMNA'26: 26th International Workshop on Computational Models of Natural Argument}

\title{Argument Structure Prediction in Online Conversations: A Comparative Study of Modeling Paradigms and Task Architectures}


\author[1,2]{Siddharth Bhargava}[%
orcid=0000-0002-2682-8557,
email=sbhargava@fbk.eu,
]
\address[1]{Fondazione Bruno Kessler, Povo 38123, Trento Italy}
\address[2]{Universidade da Coruña, 15001 A Coruña Spain}

\author[1]{Sara Tonelli}[%
orcid=0000-0001-8010-6689,
email=satonelli@fbk.eu,
]

\author[3]{Patricia Martín-Rodilla}[%
orcid=0000-0002-1540-883X,
email=p.m.rodilla@iegps.csic.es,
]
\address[3]{IEGPS-CSIC, Spanish National Research Council, 15704 Santiago de Compostela Spain}

\author[2]{Javier Parapar}[%
orcid=0000-0002-5997-8252,
email=javier.parapar@udc.es,
]


\begin{abstract}
Argument structure prediction (ASP) constructs complete argument structures from discourse by identifying argumentative units and their relations. While recent work has explored diverse approaches---including unified neural models, multi-step pipelines, and prompt-based large language models (LLMs)---their relative trade-offs remain under-explored, particularly in dialogical settings.

We present a systematic evaluation of ASP under strict schema constraints, comparing supervised fine-tuning and prompt-based LLMs across single- and multi-step task architectures, generating complete argument structures from dialogical input end-to-end. We benchmark them on three diverse dialogical corpora adapted from Inference Anchoring Theory into bipolar argument structures.  Under a shared evaluation framework, we assess predictive performance, cross-domain generalization, schema compliance, and computational efficiency. Our results show that ASP remains a challenging task, with identifying argumentative relations emerging as the primary bottleneck, largely due to the implicit and context-dependent nature of dialogical argumentation. To facilitate future research, we release our data processing pipeline and end-to-end modeling framework for computational ASP on dialogical corpora.

\end{abstract}


\maketitle

\section{Introduction}
\label{intro}

Modeling argumentative interactions in natural discourse requires capturing how arguments are proposed, challenged, and acknowledged. These interactions are formalized as \textit{argument structures}, i.e., graph-based representations that connect argumentative units through supporting and/or attacking relations. Automatically predicting such structures from raw text is a central task in Argument Mining (AM)~\cite{dore.etal_2025, sazid.mercer_2022}, commonly referred to as \textit{end-to-end Argument Mining}~\cite{morio.etal_2022a} or \textit{argument structure prediction} (henceforth referred to as ASP in this study)~\cite{santin.etal_2023} in the literature. Argument structures find application across domains including law, medicine, politics, and scientific writing (e.g., \cite{santin.etal_2023, cabessa.etal_2024, farzam.etal_2024, fromm.etal_2021}).

Designing and developing ASP requires addressing two key challenges. First, ASP is \textit{theory-driven} requiring a formal representation, grounded in argumentation theory and discourse characteristics. Second, it is \textit{data-driven}, relying on large-scale annotated corpora and supervised learning to automate structured predictions from real-world data. Together, this makes development of ASP systems both methodologically and computationally demanding. These challenges are further amplified in dialogical settings~\cite{hautli-janisz.etal_2022b, schad.etal_2024}, where noisy interactions, implicit argumentative content, and complex discourse dynamics make both formal modeling and automated prediction substantially more difficult, limiting the applicability of existing approaches.

To address these challenges, existing ASP approaches have modeled the task either as a \textit{single-step} multi-task learning problem~\cite{sazid.mercer_2022, saha.etal_2022} or as a modular \textit{multi-step} pipeline~\cite{lawrenceArgumentMiningSurvey2020}. More recently, large language models (LLMs) have been adapted for ASP through prompting or supervised fine-tuning~\cite{cabessa2024argument, gorur2025large, cruickshank.ng_2025}. These approaches differ in their modeling assumptions, training data, and evaluation, making systematic comparisons difficult.  

To systematically compare these approaches, we evaluate ASP in dialogical settings across two widely used modeling paradigms: \textbf{supervised fine-tuning} and \textbf{prompting with self-refinement}. Within each paradigm, we consider two task architectures: as a \textit{single-step} formulation and as a \textit{multi-step} ASP pipeline that first identifies argument units and then classifies the argumentative relations between them. All configurations adhere to a unified argument schema, ensuring consistent inputs and directly comparable outputs. Our proposed experimental design enables us to qualitatively study the trade-offs between modeling paradigms and task architectures, and assess their feasibility for large-scale dialogical argumentation. 

We benchmark these approaches on \textbf{three} dialogical corpora from AIFdb \cite{lawrence.etal_2012}, annotated using Inference Anchoring Theory (IAT) \cite{budzynska2011whence} which links dialogical events such as ``arguing", ``questioning", and ``agreeing'' to argumentative relations---namely inferential, conflicting, or rephrasing---between pairs of propositions identified in the text. To faciliate the computational modeling of the argument structure, we systematically convert the complex IAT annotations into bipolar argument structures, of argument units linked by either support or attack relations. Our three selected corpora vary in their interaction mode (offline vs. online), topic (political vs. non-political), and dialogue goals (collaborative vs. persuasion), enabling cross-domain evaluation.

The main contributions of this work are as follows:
\begin{itemize}
\item We introduce a systematic adaptation of IAT-based dialogical corpora into bipolar argument structures, facilitating benchmarking and computational modeling;
\item We develop an open-source, end-to-end pipeline for ASP across multiple task architectures (single-step and multi-step) and modeling paradigms (supervised fine-tuning and prompt-based generative approaches);
\item We propose a comprehensive evaluation framework for analyzing performance, generalization, schema compliance, and computational efficiency of argument structures across modeling configurations under a shared schema.
\end{itemize}

The data and system code have been released for public use (see Appendix~\ref{app:res}).

\section{Related Work}
\label{work}

ASP has been studied under different \textbf{task architectures}, primarily distinguished by their level of modularization. \textit{Multi-step (modular) approaches} decompose ASP into sequential subtasks such as argument unit identification, argument classification, and relation prediction~\cite{lawrenceArgumentMiningSurvey2020, stab.gurevych_2014a}. Relation prediction may be further divided into relation identification and relation type classification~\cite{stab.gurevych_2014a, ye.teufel_2021}. Each subtask undergoes its own optimization and training before integration.

In contrast, \textit{single-step (joint) approaches} model ASP as a unified structured prediction problem, i.e., jointly modeling argument units and relations within a single optimization and training framework. These methods leverage shared representations across subtasks, including entity-relation formulations~\cite{sazid.mercer_2022} and parsing-based or pointer architectures with dialogical extensions~\cite{saha.etal_2022, bao.etal_2022}. While they reduce error propagation and enable multi-task learning, ensuring structural validity remains challenging. Overall, multi-step approaches favor interpretability and task specialization, whereas single-step models provide integrated learning and resource efficiency at the cost of error propagation. However, their direct comparisons remain limited, especially in shared constraints.

From a \textbf{computational modeling} perspective, ASP has seen paradigm shifts from feature-based methods to training neural models and, more recently, adapting LLMs for argument mining~\cite{shi.etal_2026}. Early feature-engineered approaches~\cite{stab.gurevych_2014a, nguyen.litman_2016} offered interpretability but required substantial manual effort and lacked generalization. Neural models improved performance by learning contextual representations~\cite{lawrenceArgumentMiningSurvey2020, ye.teufel_2021}, though they depend on large annotated datasets and often struggle with out-of-distribution robustness.

Recent LLM-based approaches leverage pretrained knowledge and strong contextual reasoning to generate argument structures~\cite{cabessa2024argument, gorur2025large, cruickshank.ng_2025}. However, they may lack grounding in argumentation theory, offering limited interpretability, and may rely on superficial statistical patterns observed in data rather than robust reasoning~\cite{feger.etal_2025}. 

Prior work typically studies the modeling paradigms and task architectures as individual design choices~\cite{lawrenceArgumentMiningSurvey2020, shi.etal_2026}. We address this gap by systematically comparing fine-tuning and prompt-based approaches across both single- and multi-step architectures, evaluating their effect on performance, cross domain generalization, efficiency, and structural validity in diverse dialogical settings.

\section{Data}
\label{data}

\begin{figure}[t]
    \centering
    \includegraphics[width=\textwidth]{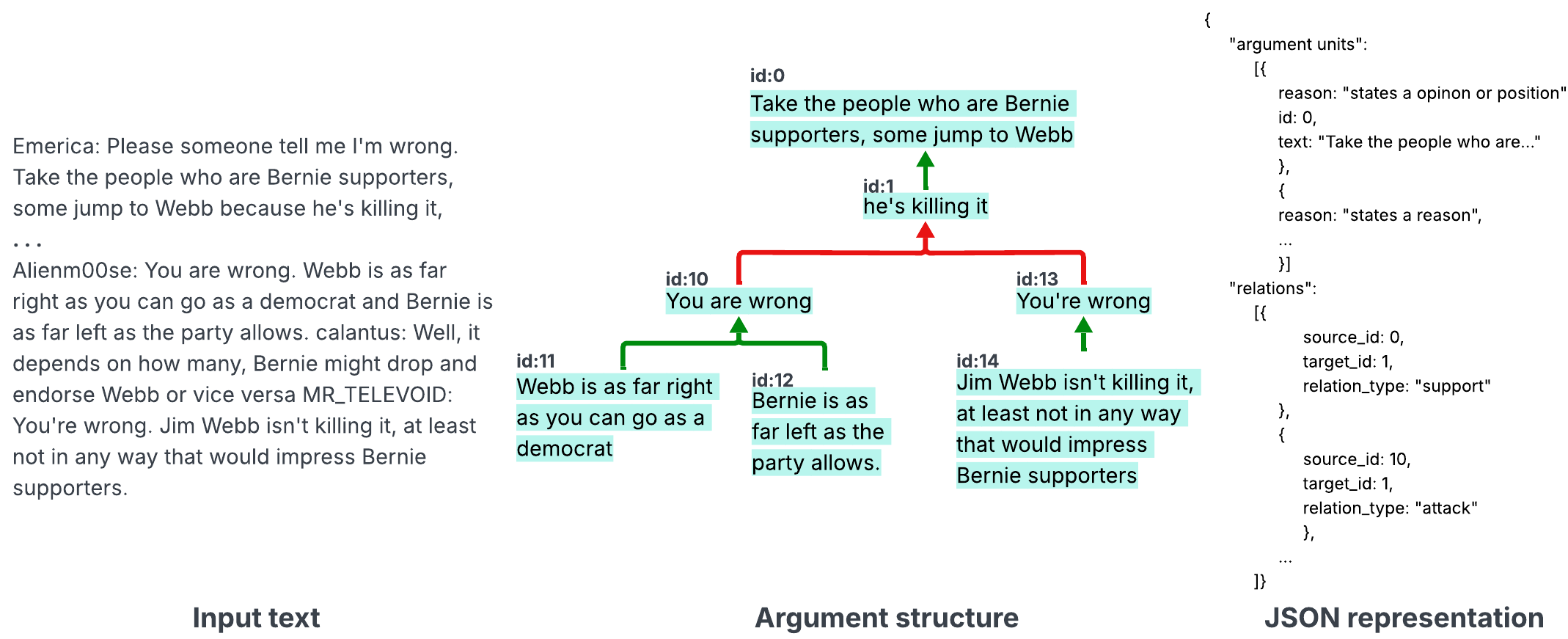}
    \caption{\textbf{Illustrative example of a conversation excerpt} [id:10055] from the US2016reddit corpus (left), the bipolar argument structure representation, annotated by humans (middle)---where green edges denote \textit{support} relations and red edges denote \textit{attack} relations---and the JSON output representation (right) indicating how the argument structure has been stored.}
    \label{fig:example}
\end{figure}

Argument structure prediction in dialogical settings remains a challenging task due to limited annotated resources available to model and train models. To mitigate this, we use AIFdb\footnote{\url{https://corpora.aifdb.org/}}, a large repository of dialogical AM corpora annotated by different AM research teams, and represented in the Argument Interchange Format (AIF) \cite{rahwanArgumentInterchangeFormat2009}. We adapt the original IAT-annotations to form simplified argument structures that contain support and attack relations (i.e., bipolar nature) between argument units extracted from the raw conversations. We refer to this representation as \textit{bipolar argument structures}, based on bipolar argumentation framework~\cite{potyka_2020}, where each structure contains a set of arguments, a set of support relations, and a set of attack relations. In this study, we focus only on IAT-annotated corpora that can be adapted to a bipolar argument structure via a common data processing framework (see Appendix~\ref{app:res}). 

From the IAT annotations, we identify all \emph{locution} nodes (speaker utterances taken verbatim from text) linked by \textbf{at least one} \emph{inferential} or \emph{conflict} relation nodes. All locution nodes satisfying this condition form the \textbf{argument units}. The resulting inferential and conflict relations are mapped to \textbf{support} and \textbf{attack} respectively, yielding a bipolar argument structure. Figure~\ref{fig:example} illustrates an example conversation with its full representation, linking the raw conversation (left), the resulting argument structure from our adaptation (middle), and the corresponding JSON output (right). All conversations have been stored in this format. 

\textbf{Argument Units.} We report in Figure~\ref{fig:example} an example of an argument structure where each unit is a span of text expressing a coherent argumentative intent, as based on the IAT framework, extracted verbatim from the conversation. All units include a unique identifier (\texttt{id}) and text (\texttt{text}) field; Additionally, a \texttt{reason} field is included that briefly describes the role of the unit. Identifiers must be unique, sequential, and chronologically ordered (starting from 0). Units must contain at least three words or form a semantically coherent span.

\begin{table*}[t]
\centering
\begin{tabular}{l r r r r r r}
\toprule
Corpus  & \#Conv. (orig.) & AvgLen & \#Units & \#Support(\%) & \#Attack(\%) & \#LongR(\%) \\
\midrule
QT30 (train) & 669 (1478) & 299.2 & 6451 & 3580 (85.9) & 565 (14.1) & 277 (6.7)\\
QT30 (test)  & 168 (1478) & 304.3 & 1559 & 871 (87.1) & 141 (12.9) & 76 (7.8) \\
US2016reddit   & 152 (264)  & 221.1 & 1666 & 895 (71.3) & 361 (28.7) & 315 (25.6) \\
RIP1    & 79 (204) & 406.3 & 765  & 386 (83.5) & 76 (16.5) & 113 (25) \\
\bottomrule
\end{tabular}
\caption{\textbf{Corpus statistics after cleaning and filtering}. Values in parentheses in the \#Conv. column denote the original number of conversations in dataset before filtering. Abbreviations: AvgLen = average conversation length (in tokens); LongR= long-distance relations where source and target are at least 3 sentences apart.}
\label{tab:data-statistics}
\end{table*}

\textbf{Argument Relations.} Relations are directed links between units labeled as \textit{support} or \textit{attack} (Figure~\ref{fig:example}). Each relation must contain a source unit ID, a target unit ID, and a relation type field. Additionally we maintain that the source--target pairs are chronologically ordered i.e., source unit must occur after target unit. This is compatible with the IAT framework.

Our data pipeline processes IAT corpora (in AIF format) to produce the bipolar argument structures. We have optional support for adding the third \textit{rephrasing} relation type, preserving speaker and dialogical act information. In this study we do not explore rephrasing relations when modeling our argument structures, and leave the investigation for future work. For more details, refer to our data processing pipeline repository (see Appendix~\ref{app:res}).

For this study, we select three corpora based on size and discourse characteristics: \textbf{QT30} (BBC Question Time debates) \cite{hautli-janisz.etal_2022b}, \textbf{US2016reddit} (Reddit reactions to the 2016 U.S. presidential debates) \cite{visserArgumentation2016US2020}, and \textbf{RIP1} (collaborative mystery game conversations) \cite{schad.etal_2024}. To support fine-tuning and in-context learning, the largest corpus (QT30) is split into training and test sets, adopting a 80:20 split with random state 42 and stratified by attack-relation frequency to preserve class distribution. Since we perform a cross-validation strategy for fine-tuning our models, we do not define an explicit validation set, as explained in Experimental Setup~\ref{sec:exp}.

To maintain data consistency and ensure some argumentative activity, we remove instances with fewer than three units or two relations and limit conversations to 1,500 tokens due to computational constraints. Table~\ref{tab:data-statistics} presents the number of instances retained following these constraints. Additional preprocessing of the discussion text and argument structure is done to remove non-ASCII characters, markup, duplicates, nested units, and invalid self-relations.

QT30 and US2016reddit contain persuasive political discourse, whereas RIP1 captures collaborative, non-political interaction. As shown in Table~\ref{tab:data-statistics}, US2016reddit contains shorter units and a higher proportion of attacks (approximately 2.5 per conversation) compared to the offline corpora. Overall, attack relations remain relatively sparse across datasets (approximately 1.5 attacks versus 6 supports per conversation). Both US2016reddit and RIP1 exhibit a higher proportion of long-distance relations, while QT30 remains largely local. Although all corpora follow the IAT framework, minor variations arise from differences in annotation practices across research groups. All datasets have been used in accordance with their original licenses and ethical guidelines.


\section{Models}
\label{sec:models}

We investigate two modeling paradigms for ASP: \textbf{fine-tuning} of deep learning models and \textbf{prompting} large language models. Both are evaluated under two task architectures: a \textbf{single-step} formulation, where we adopt a multi-task learning approach and directly produce the argument structure, and a \textbf{multi-step} formulation, where the tasks are executed sequentially with intermediate, schema-validated outputs. Table~\ref{tab:models} summarizes the models used in this study and introduces the model codes used to refer to the respective configurations in this paper.

\begin{table}[t]
\centering
\begin{tabular}{l l l l}
\toprule
Paradigm & Model & Arch. & Code (MC) \\
\midrule
Fine-tuning 
& BiLSTM-ER & ss & FT \\
& XLM-R-Long + XLM-R-Large & ms & FT \\
\midrule
Prompting
& Qwen2.5-14B-Instruct & ss / ms & Q0, Q3, Q5 \\
& DeepSeek-R1-Distill-Qwen-14B & ss / ms & D0, D3, D5 \\
& Gemma3-12B-it & ss / ms & G0, G3, G5 \\
\bottomrule
\end{tabular}
\caption{Models and configurations evaluated. Paradigms include supervised fine-tuning and prompt-based inference. Task Architectures: single-step (ss) and multi-step (ms). Model codes (MC) denote the model and number of context examples (e.g., Q3 = Qwen with three examples).}
\label{tab:models}
\end{table}

\subsection{Supervised fine-tuning}
\label{sec:ftapp}
For the \textbf{single-step} fine-tuning, we adopt the framework proposed in \cite{sazid.mercer_2022}, which employs a biLSTM-ER (Entity–Relation) model to perform joint token-level prediction for both unit and relation. The model simultaneously predicts (i) argumentative spans via BIO tagging, (ii) relation types (support or attack), and (iii) relational distance estimates (ranging from $-8$ to $+8$ sentences) indicating the relative position of the target unit. 

For the \textbf{multi-step} fine-tuning, we train two models independently for argument unit extraction (AUE) and relation type classification (RTC). The AUE component performs BIO-tagged span-identification using XLM-RoBERTa-Longformer~\cite{Sagen1545786} to identify argumentative spans within conversation excerpts with max length 1500. The RTC component performs sequence-level classification using XLM-RoBERTa-large~\cite{DBLP:journals/corr/abs-1911-02116} that takes as input a pair of predicted argument units, paired chronologically (i.e., source is always after target) and provided with a limited context window of 150, and determines the relation type from the set $\{\textit{support},\textit{attack},\textit{no-relation}\}$. To improve discrimination between relational and non-relational pairs, RTC training data is augmented with a proportional number of negative samples, including invalid unit–unit and unit–non-unit pairs.

\subsection{Prompting with self-refinement}
\label{sec:llmapp}

We employ in-context learning with three widely used open-source, medium-sized LLMs: \textbf{Qwen2.5-14B-Instruct}~\cite{qwen.etal_2025}, \textbf{DeepSeek-R1-Distill-Qwen-14B}~\cite{guo.etal_2025}, and \textbf{Gemma-3-12b-it}~\cite{team.etal_2025}, accessed via Hugging Face. Prompts are designed for both \textit{single-step} and \textit{multi-step} settings using meta-prompting techniques~\cite{hou-etal-2022-metaprompting}, with argument schema constraints embedded in the system message~(see Appendix~\ref{app:prompt}). Final templates are selected empirically and adapted to each model’s chat-completion format. In the \textbf{single-step} setting, models generate full argument structures in one pass, whereas in the \textbf{multi-step} setting, the process is decomposed into unit extraction followed by relation prediction (\textit{support}/\textit{attack}) between the predicted units.

To enforce schema compliance, outputs are validated using Pydantic~\cite{pydantic}. Invalid outputs trigger a single retry, where validation errors and corrective instructions are injected into the prompt automatically, following self-refinement strategies for reducing hallucinations and incomplete outputs~\cite{madaan.etal_2023}.

To study supervision effects, we vary in-context examples (zero-, three-, and five-shot), sampled from QT30 training data with balanced support and attack coverage. Examples include brief explanations of each unit’s dialogical role, inspired by speech-act information in IAT annotations (Figure~\ref{fig:example}). Models are prompted to generate such explanations to support more grounded unit and relation predictions.


\section{Experimental Setup}
\label{sec:exp}
The experimental setup describes training and inference for both fine-tuned and prompting approaches (Table~\ref{tab:models}). All experiments have been conducted on a single 48GB NVIDIA Ampere A40 GPU.

\textbf{Fine-tuning Language Models.}
The single- and multi-step models were fine-tuned using the QT30 training corpus. Hyperparameter optimization was performed using Optuna \cite{optuna2019}, with eight trials per configuration to select the learning rate from a log-uniform range of $[1\mathrm{e}{-5}, 5\mathrm{e}{-4}]$. Batch size and dropout were fixed to 8 and 0.1. To address label imbalance across subtasks, weighted loss functions were applied. Using the best hyperparameters, models were then re-trained using 5-fold cross-validation with early stopping based on evaluation loss (patience = 3), ensuring a stable estimate of performance and a robust model selection criterion. 

\textbf{Inference with Finetuned Models.}  
Predictions were generated using the best-performing models and post-processed according to the schema validation logic (Section~\ref{data}). In single-step, this involved correcting BIO predictions to resolve span inconsistencies and relational distance conflicts. In the multi-step, malformed spans (those containing an errored word or only punctuation) were filtered and chronologically valid unit pairs were constructed before relation prediction. The final argument structure consists of schema-compliant units with validated support and attack relations.

\textbf{Inference with Prompted LLMs.} 
LLM inference was performed using the Outlines Python library \cite{willard2023efficient}, which enables structured JSON generation and integrates with our Pydantic-based schema validation (Section~\ref{sec:llmapp}). Decoding parameters were fixed to temperature $0$, top\_p $0.0$, random seed $42$, and repetition penalty $1.0$ to ensure deterministic outputs and improve schema adherence. Maximum \textbf{two} tries per conversation are allowed.


\section{Results}
\label{results}
A prediction is considered a valid generation if it satisfies the schema (see Section~\ref{data}), thereby avoiding malformed/empty outputs. Minor violations (e.g., single-word units or invalid IDs) are corrected via post-processing, while major inconsistencies in schema result in invalid predictions.

All models are evaluated on three datasets: QT30-test (in-domain), US2016reddit (cross-domain), and RIP1 (cross-domain). We assess predictive performance, cross-domain generalization, computational efficiency, and schema compliance.

We follow the evaluation protocol of \cite{sazid.mercer_2022}, which decomposes ASP into two phases: argument unit evaluation and relation evaluation.

Argument units are evaluated at the span level by aligning predicted and gold spans using a one-to-one best-alignment procedure under two span overlap thresholds: 50\% partial overlap and 100\% exact match~\cite{sazid.mercer_2022}. A predicted span is considered correct if it is aligned to a gold span whose overlap exceeds the threshold. We measure precision, recall, and F1 for the \texttt{ARG} and \texttt{NON-ARG} class, counting unaligned predicted spans as false positives and unaligned gold spans as false negatives.

Relation evaluation begins with first assessing whether aligned units are paired with their correct counterparts. For correctly aligned and paired units, we then evaluate whether the predicted relation type (\texttt{support} or \texttt{attack}) between them is correct. A relation is counted as a true positive if (i) the  source and target of predicted pair are aligned, (ii) the predicted pair also exists in the gold structure, and (iii) the relation type is correct, otherwise noted as false . Relation-level precision, recall, and F1 are measured under both span-overlap thresholds.

\subsection{Task Performance and Generalization}

\begin{figure*}[t]
\centering

\begin{subfigure}[t]{0.48\textwidth}
    \centering
    \includegraphics[height=0.30\textheight, angle=-90]{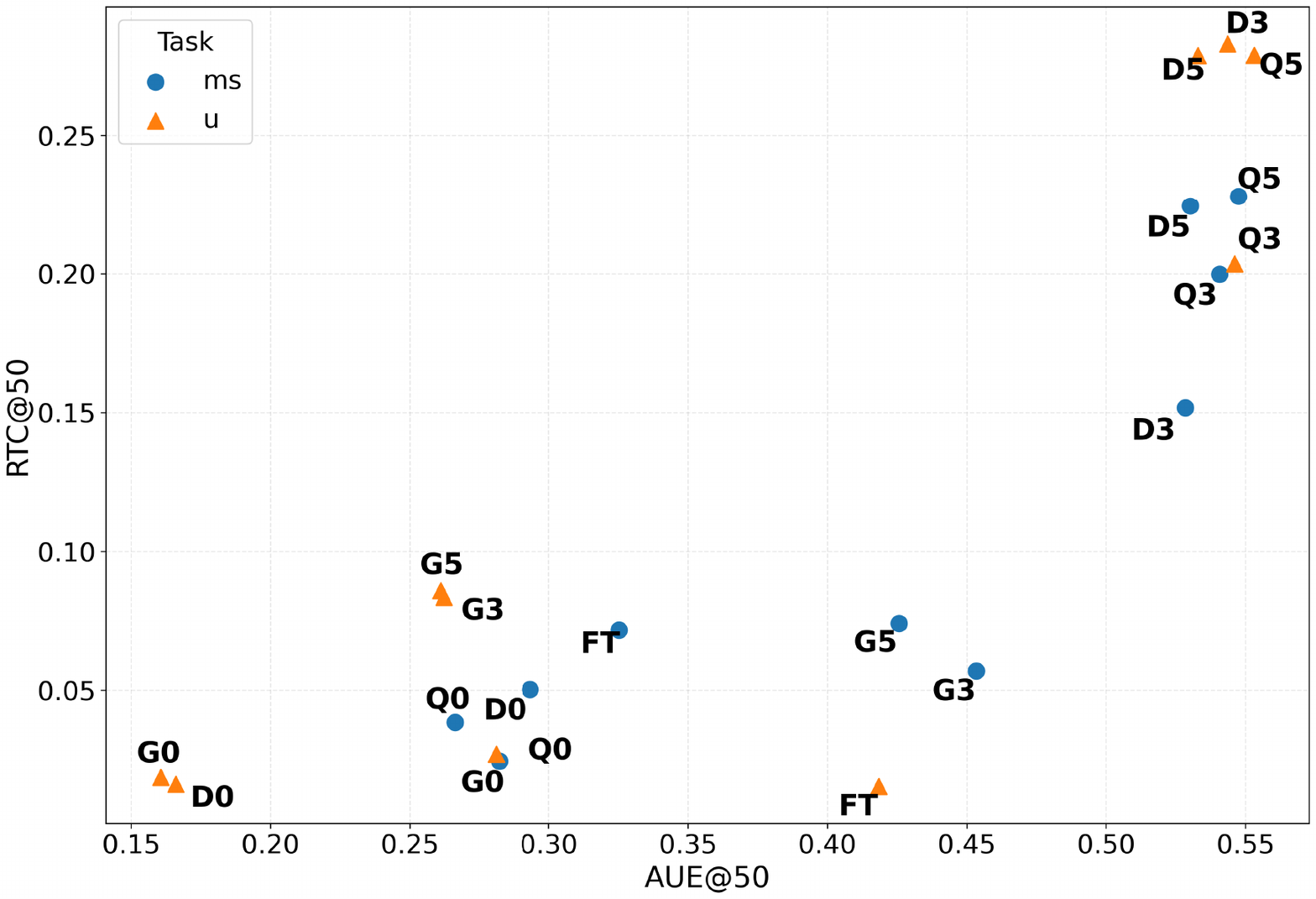}
    \caption{QT30-test}
\end{subfigure}
\hspace{0.02\textwidth}
\begin{subfigure}[t]{0.48\textwidth}
    \centering
    \includegraphics[height=0.30\textheight, angle=-90]{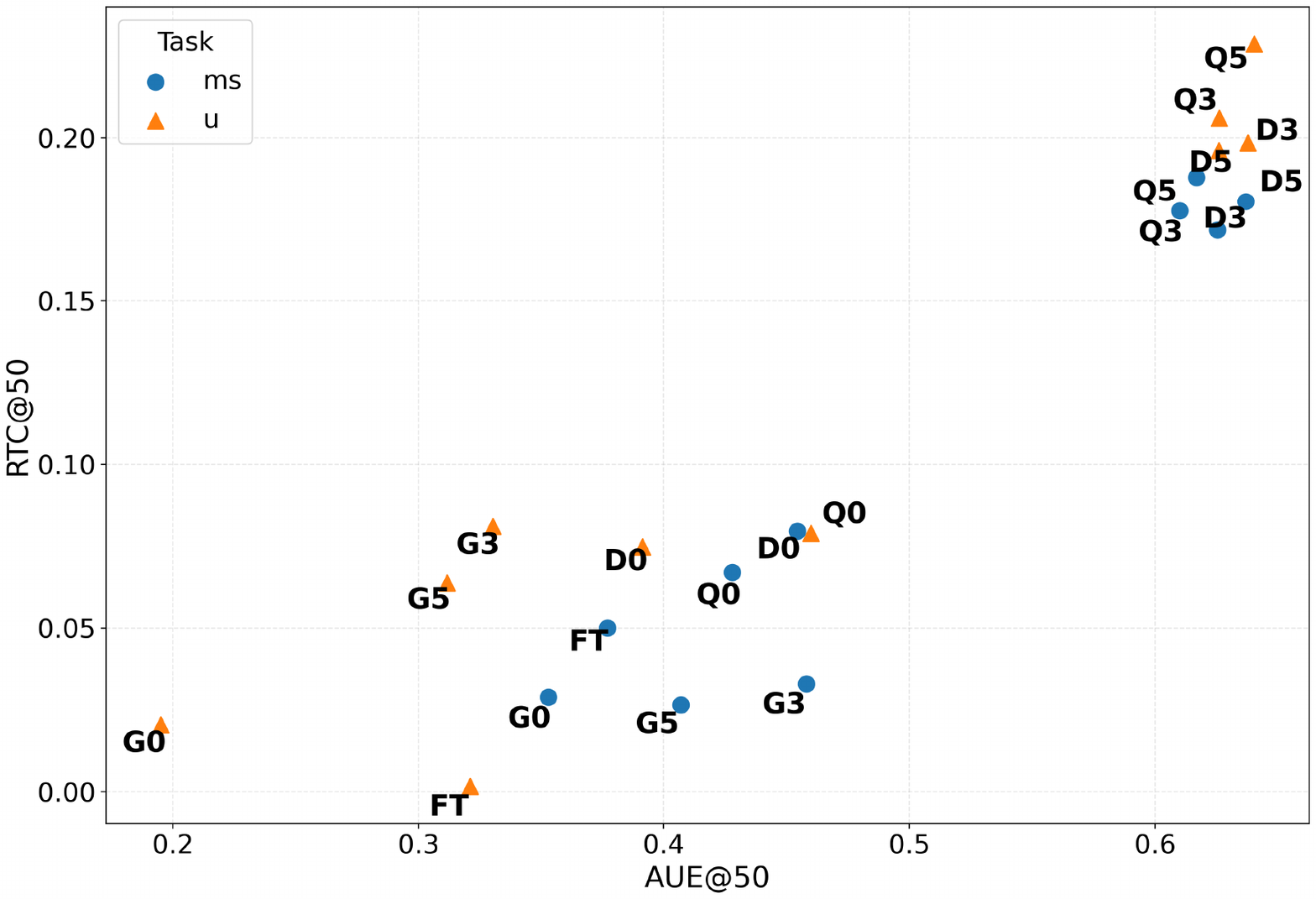}
    \caption{US2016reddit}
\end{subfigure}

\begin{subfigure}[t]{0.45\textwidth}
    \centering
    \includegraphics[height=0.30\textheight, angle=-90]{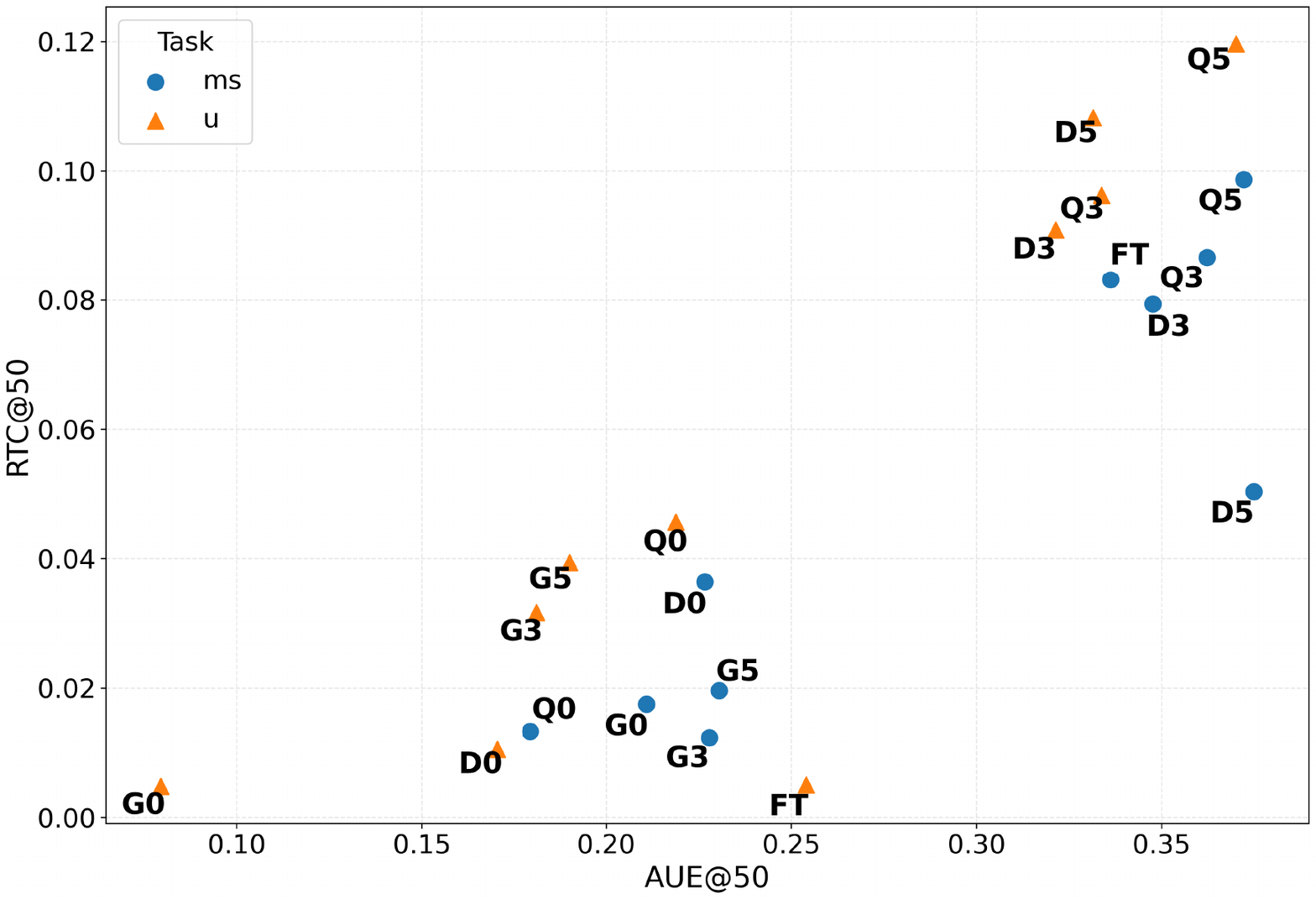}
    \caption{RIP1}
\end{subfigure}

\caption{
\textbf{Task performance across datasets and configurations} (identified by Model Codes [MC]; see Table~\ref{tab:models}) for single- (\textit{ss}) and multi-step (\textit{ms}) settings. \textit{Plots (a–c)} show argument unit extraction (x: AUE@50) and relation classification (y: RTC@50) at 50\% span overlap, with each point representing the weighted F1 score (AUE, RTC).
}
\label{fig:aue-rtc-scatter}
\end{figure*}

To analyze model and architectural influence on ASP performance, we present corpus-wise scatter plots in Figure~\ref{fig:aue-rtc-scatter}. The x-axis represents weighted F1 for argument unit extraction (AUE) subtask, while the y-axis represents weighted F1 score for relation type classification (RTC) subtask, both evaluated under the 50\% span-overlap condition.

In-domain results on QT30-test (Figure~\ref{fig:aue-rtc-scatter}a) show that single-step architecture achieve stronger RTC performance than multi-step architecture though they perform similarly on AUE. Relation prediction benefits from the joint modeling.

Prompt-based models with more context examples (Q${3,5}$, D${3,5}$) consistently outperform fine-tuned models, while zero-shot and Gemma-based configurations perform the poorest. This indicates that adding context examples facilitates both the subtasks significantly. However, this behaviour is not uniform across the different LLM families. Qwen2.5-14B-Instruct demonstrated the best performance due to better schema compliance while Gemma3-12B-it suffered from poor schema compliance that resulted in malformed generations. We discuss this further in the next section.

Studying the cross-domain results over the other two corpora, US2016reddit and RIP1 (Figure~\ref{fig:aue-rtc-scatter}b–c) shows a significant performance drop for RTC than AUE indicating that relations do not generalize well between the different corpora. Single-step architectures continue to outperform the multi-step for RTC over AUE indicating that RTC does benefit from the joint modeling. Performance is higher on US2016reddit than on RIP1, likely due to US2016reddit having closer alignment with QT30 in annotation and discourse characteristics. RIP1 conversations are generally longer, have sparser distribution of units and relations, and are non-political and collaborative in nature. These distinct characteristic differences likely resulted in the significant performance drop compared to the other two corpora.

Our findings show that prompt-based single-step configurations performed the best among all the tested configurations. However, none of the configurations generalized well to unseen corpora or different dialogical setting. 

\subsection{Computational Efficiency and Schema Compliance}
\label{sec:util}

While prompting generally outperforms fine-tuning, it incurs higher computational cost and more frequent schema violations (malformed/empty outputs), likely due to hallucinated outputs and the difficulty of generating complete schema-compliant outputs with reasoning enabled.

\begin{table*}[t]
\centering
\begin{tabular}{lccc ccc}
\toprule
& \multicolumn{3}{c}{\textbf{QT30-test} (168 conv.)} 
& \multicolumn{3}{c}{\textbf{RIP1} (79 conv.)} \\
\cmidrule(lr){2-4} \cmidrule(lr){5-7}
MC & SR(\%) & T(s) & Tk 
   & SR(\%) & T(s) & Tk \\
\midrule

Q0$_{ss}$ & 96.4$_{+3.6}$ & 27.46$_{+0.6}$ & 210k$_{+13k}$ 
          & 100.0$_{+0.0}$ & 37.38$_{+0.0}$ & 119k$_{+0k}$ \\
Q3$_{ss}$ & 100.0$_{+0.0}$ & 34.49$_{+0.0}$ & 623k$_{+0}$ 
          & 98.7$_{+1.3}$ & 34.45$_{+0.0}$ & 301k$_{+8k}$ \\
Q5$_{ss}$ & 100.0$_{+0.0}$ & 37.89$_{+0.0}$ & 830k$_{+0}$ 
          & 98.7$_{+1.3}$ & 37.3$_{+0.3}$ & 397k$_{+10k}$ \\
D3$_{ss}$ & 100.0$_{+0.0}$ & 46.11$_{+0.0}$ & 646k$_{+0}$ 
          & 97.4$_{+0.0}$ & 48.78$_{+2.0}$ & 312k$_{+4k}$ \\
D5$_{ss}$ & 100.0$_{+0.0}$ & 47.76$_{+0.0}$ & 847k$_{+0}$ 
           & 97.4$_{+0.0}$ & 52.18$_{+2.0}$ & 405k$_{+4k}$ \\
G5$_{ss}$ & 35.7$_{+22.0}$ & 132.46$_{+90}$ & 529k$_{+529k}$ 
          & 19.0$_{+20.3}$ & 149.7$_{+120}$ & 210k$_{+270k}$ \\

\midrule
Q0$_{ms}$ & 98.8$_{+1.2}$ & 18.6$_{+0.0}$ & 284k$_{+4k}$ 
          & 98.7$_{+1.3}$ & 16.6$_{+0.0}$ & 147k$_{+2k}$ \\
Q3$_{ms}$ & 99.4$_{+0.6}$ & 32.55$_{+0.0}$ & 1049k$_{+8k}$ 
          & 100.0$_{+0.0}$ & 27.14$_{+0.0}$ & 503k$_{+0}$ \\
Q5$_{ms}$ & 100.0$_{+0.0}$ & 35.27$_{+0.0}$ & 1416k$_{+0}$ 
          & 100.0$_{+0.0}$ & 31.02$_{+0.0}$ & 676k$_{+0}$ \\
D3$_{ms}$ & 95.8$_{+4.2}$ & 42.57$_{+1.0}$ & 1069k$_{+49k}$ 
          & 79.74$_{+18.98}$ & 38.04$_{+2.5}$ & 493k$_{+78k}$ \\
D5$_{ms}$ & 50.0$_{+50.0}$ & 40.27$_{+3.0}$ & 1203k$_{+263k}$ 
          & 15.2$_{+78.5}$ & 32.43$_{+5.0}$ & 476k$_{+262k}$ \\
G5$_{ms}$ & 11.9$_{+80.9}$ & 111.58$_{+63.0}$ & 948k$_{+621k}$ 
          & 12.6$_{+78.5}$ & 118.01$_{+60.0}$ & 448k$_{+349k}$ \\
\bottomrule
\end{tabular}
\caption{
\textbf{Computational efficiency of configurations across QT30-test and RIP1}. SR: success rate of schema-compliant predictions in first try; T: Average time taken per conversation in first try; Tk: total tokens (in thousands) outputted for all conversations in first try. Subscripts (+x) denote changes observed when a retry is performed.
}
\label{tab:eff}
\end{table*}

Minor violations, such as poorly segmented units (1–2 words or invalid IDs), are corrected through post-processing, while major violations (e.g., incomplete JSON structures) trigger a single retry via self-refinement. In this step, the model is re-prompted with error reports and corrective feedback. While this improves schema compliance, it increases runtime and token usage. Table~\ref{tab:eff} reports first-pass success rates for generating schema-compliant structures, along with average inference time and total token consumption, and quantifies the additional cost introduced by retries.

Our findings show that effectiveness of the retry mechanism is strongly dependent on task architecture: multi-step settings benefit more, whereas single-step already achieve high first-try performance and exhibit only marginal gains if retry was triggered. Analysis of multi-step predictions indicates that retries are triggered more frequently during the argument unit extraction (AUE) stage, largely due to over-prediction of units. In contrast, single-step architectures may benefit from more grounded predictions through joint unit–relation modeling.

Retries are most beneficial in low-context settings (e.g., zero-shot), where they yield higher gains per additional token. However, they substantially increase computational cost---often nearly doubling token usage---while providing limited overall improvements.

Overall, these results highlight a trade-off between generating complex structured outputs and maintaining computational efficiency. Future work should explore selective retry strategies, such as triggering retries based on confidence estimates or restricting them to low-context settings, to better balance this trade-off.

\subsection{Error Analysis}
\label{sec:error}

We conducted an error analysis of the model predictions for our best performing configurations and the three corpora. We study three main stages in ASP here: the main failure modes in AUE subtask, and the RTC subtask split into relation prediction (source-target unit pairing) and type classification (support--attack).

Analyzing the predicted argument units obtained from AUE, the primary errors occurred due to segmentation inconsistencies, i.e., the models failed to segment and extract units from the raw text efficiently. More often, the LLM-based models tend to over-segment the text resulting in poor precision. Adopting a 50\% partial overlap allowed for better unit alignment with the gold units and resulted in better recall (Table~\ref{tab:errors}). However, at 100\% perfect overlap, the performance drastically falls indicating that the models struggle to efficiently identify segment boundaries. Explicit segmentation with clear rules is needed for improving performance in argument unit prediction and extraction, especially in dialogical settings where statements are often incomplete or poorly-constructed making argument boundaries difficult to be assessed.

Analysing the downstream relation prediction, one of the main failure modes is the partially-overlapped units getting incorrectly paired as source-targets. This led to \textit{endpoint misalignment} with the gold pairs (when either the source or target fails to align with a gold unit) and made the relation in-valid. Additionally, most LLM-based approaches behaved conservatively and predicted few relations, that led to higher number of missing relations. These two errors resulted in significantly poor recall as most relations were missing or marked missing due to in-validity for evaluation (Table~\ref{tab:errors}). Since only correctly paired relations proceeded for evaluation, precision was relatively stable.

\begin{table*}[t]
\centering
\begin{tabular}{lcccccc}
\toprule
& \multicolumn{2}{c}{\textbf{QT30-test}} 
& \multicolumn{2}{c}{\textbf{US2016reddit}} 
& \multicolumn{2}{c}{\textbf{RIP1}} \\
\cmidrule(lr){2-3} \cmidrule(lr){4-5} \cmidrule(lr){6-7}
MC & AUE$_{P/R}$ & RTC$_{P/R}$ 
   & AUE$_{P/R}$ & RTC$_{P/R}$ 
   & AUE$_{P/R}$ & RTC$_{P/R}$ \\
\midrule

FT$_{ms}$ 
& 0.32\,/\,0.35 & 0.12\,/\,0.06 
& 0.42\,/\,0.36 & 0.08\,/\,0.04 
& 0.28\,/\,\textbf{0.44} & 0.12\,/\,0.08 \\

Q0$_{ss}$ 
& 0.35\,/\,0.25 & 0.06\,/\,0.02 
& 0.55\,/\,0.42 & 0.18\,/\,0.05 
& 0.21\,/\,0.25 & 0.12\,/\,0.03 \\

Q3$_{ss}$ 
& 0.52\,/\,0.61 & 0.33\,/\,0.16 
& 0.64\,/\,0.64 & 0.35\,/\,0.16 
& 0.32\,/\,0.38 & 0.25\,/\,0.06 \\

Q5$_{ms}$ 
& \textbf{0.53}\,/\,0.61 & 0.38\,/\,0.18 
& \textbf{0.67}\,/\,0.60 & 0.30\,/\,0.14 
& \textbf{0.38}\,/\,0.38 & 0.24\,/\,0.06 \\

\textbf{Q5$_{ss}$} 
& 0.51\,/\,\textbf{0.64} & \textbf{0.44}\,/\,0.22 
& 0.65\,/\,0.65 & \textbf{0.41}\,/\,\textbf{0.17} 
& 0.34\,/\,0.43 & \textbf{0.27}\,/\,\textbf{0.08} \\

D5$_{ms}$ 
& 0.46\,/\,0.68 & 0.32\,/\,0.19 
& 0.62\,/\,\textbf{0.69} & 0.25\,/\,0.15 
& 0.30\,/\,0.40 & 0.19\,/\, 0.06 \\

D5$_{ss}$ 
& 0.45\,/\,0.69 & 0.40\,/\,\textbf{0.24} 
& 0.59\,/\,0.69 & 0.29\,/\,0.14 
& 0.28\,/\,\textbf{0.44} & 0.21\,/\,0.08 \\

\bottomrule
\end{tabular}

\caption{
\textbf{Performance analysis across model configurations} (MC, see Table~\ref{tab:models}). Precision (P) and recall (R) are reported as P\,/\,R for argument unit extraction (AUE) and relation classification (RTC). Bold values indicate the best performance within each dataset.
}
\label{tab:errors}
\end{table*}

Through manual inspection, two factors are identified: (i) limited contextual understanding and (ii) long-distance dependencies. Operating in a multi-party dialogical settings that contain implicit and noisy instances, models often failed to accurately resolve the implicitness in the discussion based on the limited context provided. For instance, linking ``You are wrong” to ``he's killing it” requires broader dialogical context from subsequent turns (see Figure~\ref{fig:example}). Single-step configurations, especially prompt-based configurations, may be less susceptible to this since they model the whole conversation directly. More work is needed to investigate this further. Additionally, in RIP1 corpus and US2016reddit corpus, that have relatively longer discussions and long-distance argumentative interactions (Table~\ref{tab:data-statistics}), configurations failed to identify these connections and mostly predicted relations between immediate neighbours.

Finally, analyzing the correctly paired relations for type classification, most configurations performed well, achieving near-perfect \textit{support} and good \textit{attack} scores despite the inherent class imbalance.  This indicates that relation pairing, rather than type classification, is the primary bottleneck. More details on the quantitative error analysis is provided in Appendix~\ref{app:qua-err}. Future work must decouple the relation prediction and classification type, i.e., link argument units as potential source--target pairs explicitly before predicting the nature of the relation itself.


\section{Conclusion}
\label{conc}

In this work, we address ASP for online conversations. We adapt diverse IAT-annotated corpora into bipolar argument structures enabling their use for computational ASP. Building on this representation, we conduct a systematic evaluation of ASP, comparing supervised fine-tuning and prompt-based LLMs across single- and multi-step task architectures. Our analysis examines the effect of model design choices on performance, generalization, schema compliance, and computational efficiency under shared structural constraints. We release our data processing and modeling pipelines, which supports IAT-based corpora in AIF format and can be extended to a wide range of language models for dialogical argumentation (Appendix~\ref{app:res}).

Our findings find that predicting argument structures from online conversations remains a challenging task. Testing various modeling configurations over diverse dialogical datasets demonstrates that the key bottleneck in ASP remains in identifying if two given argument units are argumentatively linked or not. This challenge is further amplified in dialogical settings that contain implicit or context-dependent argumentative content. It requires broader discourse context and deeper reasoning to infer argumentative relations accurately. Our results indicate the need for explicit contextual grounding when designing ASP systems for complex dialogical interactions.

A limitation of our work has been the limited modeling strategies that have been explored. Future work should extend this analysis to include fine-tuning LLMs~\cite{efeoglu.paschke_2025}, testing graph-based neural networks~\cite{sun.etal_2024a}, and adopting neuro-symbolic frameworks for contextual grounding of both arguments and their relations~\cite{plenz.etal_2024}. Another limitation is the limited cross-domain generalization of the evaluated ASP approaches, as performance degrades on datasets with distinct dialogical and argumentative characteristics. Moreover, some of the datasets used in this study may have appeared in the pre-training corpora of the evaluated LLMs. Future work should evaluate ASP on a broader and more diverse set of dialogical corpora to improve robustness to out-of-distribution data and better assess model generalization. Given the limited availability of annotated resources for dialogical settings, we hope that our data processing pipeline will facilitate future research and advance the computational modeling of dialogical ASP.

Finally, one of the main challenges to dialogical ASP is relation prediction, specifically identifying the source--target pair. Future research should focus on enhancing contextual reasoning such as supplementing contextual knowledge to an argument using retrieval augmented generation (RAG)~\cite{sun.etal_2026}, or incorporating structural knowledge to improve relation prediction directly~\cite{dore.etal_2025}. 

Developing robust and reliable dialogical ASP systems can support real-world applications that require argumentation comprehension such as in conflict resolution, collaborative reasoning and decision-making, and argument-based summarization and explanations. 


\section{Acknowledgments}

This research work has received funding from the European Union's Horizon Europe research and innovation programme under the Marie Skłodowska-Curie Grant Agreement No. 101073351. Views and opinions expressed are however those of the author(s) only and do not necessarily reflect those of the European Union or European Research Executive Agency (REA). Neither the European Union nor the granting authority can be held responsible for them.





\section*{Declaration on Generative AI}
During the preparation of this work,we used ChatGPT in order to assist with grammar, formal tone, and spelling check. We have carefully reviewed and edited the content as needed and take full responsibility for the publication’s content.

\bibliography{sample-ceur}

\appendix

\section{Online Resources}
\label{app:res}

The data and system code are available via
\begin{itemize}
    \item \href{https://github.com/The-obsrvr/ArgStrPrediction}{Github: System repository}
    \item \href{https://github.com/The-obsrvr/IAT-BAS-Data-Pipeline}{Github: Data Processing Pipeline Repository}
    \item \href{https://zenodo.org/records/19627671}{Data corpora: IAT2BAS Dataset}
\end{itemize}

\section{Prompts used for ASP}
\label{app:prompt}

We provide the templates for the three prompts used in this study here in Figures~\ref{fig:prompt_single_step},~\ref{fig:prompt_argument_units}, and~\ref{fig:prompt_relations}. More details on their use can be viewed in the Github repository.

\begin{figure}[t]
\centering
\begin{minipage}{0.95\columnwidth}
\hrule
\vspace{0.6em}
\small
\raggedright

\textbf{Prompt for Single-Step ASP}

\vspace{0.6em}

\ttfamily
You are an argument mining expert. From the discussion, extract argument units and their relations.

\vspace{0.6em}

Return ONE JSON object with this exact shape:

\vspace{0.4em}

\{
  "argument\_units": [
    \{"reason": "...", "id": 0, "text": "..."\},
    ...
  ],
  "relations": [
    \{"source\_id": 1, "target\_id": 0, "type": "support"\},
    ...
  ]
\}

\vspace{0.6em}

Argument units:
\begin{itemize}\itemsep0pt
    \item \texttt{text}: copy verbatim from the discussion; do not paraphrase.
    \item Each unit is a single argumentative idea, e.g., asserts, questions, rejects, accepts, defends, or challenges a topic or another statement.
    \item Assign IDs in order of appearance: 0, 1, 2, \ldots
    \item \texttt{reason}: short explanation of the unit's intent or expressed idea.
\end{itemize}

Relations:
\begin{itemize}\itemsep0pt
    \item Only create a relation if there is a clear indication in the conversation text.
    \item Only treat a pair as ``no relation'' if they are clearly unrelated or only share a topic without one supporting or attacking the other.
    \item \texttt{support}: source accepts or gives reasons, evidence, or clarification for target.
    \item \texttt{attack}: source challenges, rejects, undercuts, or undermines target.
    \item Only create a relation if there is a clear, explicit link, e.g., ``because'', ``but'', ``however'', or ``in response to''.
    \item If unsure or only loosely related by topic, do not create a relation.
    \item Use ONLY IDs from \texttt{argument\_units}; never invent new IDs.
    \item By default, the source should appear after the target in the discussion.
    \item Do NOT create symmetric duplicates.
\end{itemize}

Schema constraints:
\begin{itemize}\itemsep0pt
    \item Root object MUST have ONLY: \texttt{"argument\_units"} and \texttt{"relations"}.
    \item \texttt{"argument\_units"} and \texttt{"relations"} MUST each be lists.
    \item Elements of \texttt{"argument\_units"} MUST have ONLY: \texttt{"reason"}, \texttt{"id"}, and \texttt{"text"}.
    \item Elements of \texttt{"relations"} MUST have ONLY: \texttt{"source\_id"}, \texttt{"target\_id"}, and \texttt{"type"}.
    \item \texttt{"type"} MUST be exactly \texttt{"support"} or \texttt{"attack"}.
    \item Do NOT output any text before or after the JSON.
\end{itemize}

\vspace{0.6em}

Few-shot examples:

\vspace{0.3em}

[FEW-SHOT EXAMPLE 1]

[FEW-SHOT EXAMPLE 2]

\vspace{0.6em}

User input:

\vspace{0.3em}

Discussion:

<conversation>

\vspace{0.6em}
\rmfamily
\hrule
\end{minipage}
\caption{Prompt template used for single-step argument structure prediction.}
\label{fig:prompt_single_step}
\end{figure}

\begin{figure}[t]
\centering
\begin{minipage}{0.95\columnwidth}
\hrule
\vspace{0.6em}
\small

\textbf{Prompt for Argument Unit Extraction in Multi-Step ASP}

\vspace{0.6em}

\ttfamily
You are an argument mining expert. Extract argument units from the discussion.

\vspace{0.6em}

Return ONE JSON object with this exact shape:

\vspace{0.4em}

\{
  "argument\_units": [
    \{"reason": "...", "id": 0, "text": "..."\},
    ...
  ]
\}

\vspace{0.6em}

Rules:
\begin{itemize}\itemsep0pt
    \item Copy \texttt{text} verbatim from the discussion; do not paraphrase.
    \item Each unit MUST be a single argumentative idea, e.g., asserts, questions, rejects, accepts, defends, or challenges a topic or another statement.
    \item Assign IDs in order of appearance: 0, 1, 2, \ldots
    \item \texttt{reason} is a short explanation of the unit's intent or role, e.g., claim, premise, opinion, stance, or counterclaim.
    \item Exclude non-argumentative chatter, jokes, greetings, or pure rhetoric.
\end{itemize}

Schema constraints:
\begin{itemize}\itemsep0pt
    \item The root object MUST have ONLY the key \texttt{"argument\_units"}.
    \item Each conversation should return at least 3 units.
    \item Each element MUST have ONLY \texttt{"id"}, \texttt{"text"}, and \texttt{"reason"}.
    \item Do NOT output any text before or after the JSON.
\end{itemize}

\vspace{0.6em}

Few-shot examples:

\vspace{0.3em}

[FEW-SHOT EXAMPLE 1]

[FEW-SHOT EXAMPLE 2]

\vspace{0.6em}

User input:

<conversation>

\vspace{0.6em}
\rmfamily
\hrule
\end{minipage}
\caption{Prompt template used for argument unit extraction.}
\label{fig:prompt_argument_units}
\end{figure}

\begin{figure}[t]
\centering
\begin{minipage}{0.95\columnwidth}
\hrule
\vspace{0.6em}
\small
\raggedright

\textbf{Prompt for Relation Prediction and Type Classification in Multi-Step ASP}

\vspace{0.6em}

\ttfamily
You are an argument mining expert. Identify directional relations between the given units.

\vspace{0.6em}

Return ONE JSON object with this exact shape:

\vspace{0.4em}

\{
  "relations": [
    \{"source\_id": ..., "target\_id": ..., "type": "..."\},
    ...
  ]
\}

\vspace{0.6em}

Meaning:
\begin{itemize}\itemsep0pt
    \item \texttt{support}: source gives reasons, evidence, or clarification for target.
    \item \texttt{attack}: source challenges, rejects, undercuts, or undermines target.
\end{itemize}

Rules to avoid spurious relations:
\begin{itemize}\itemsep0pt
    \item Only create a relation if there is a clear indication in the text.
    \item Only treat a pair as ``no relation'' if they are clearly unrelated or only share a topic without one supporting or attacking the other.
    \item Only create a relation if the text shows a clear, explicit link to its target unit.
    \item If unsure or only loosely related by topic, do NOT create a relation.
    \item Ensure that the conversation text is considered as additional context, not only the extracted units.
\end{itemize}

Constraints:
\begin{itemize}\itemsep0pt
    \item Use ONLY IDs from the provided units; never invent new IDs.
    \item By default, the source should appear after the target in the discussion.
    \item \texttt{type} MUST be exactly \texttt{"support"} or \texttt{"attack"}; no other labels are allowed.
    \item The root object MUST have ONLY the key \texttt{"relations"}.
    \item \texttt{"relations"} MUST be a list.
    \item Do NOT output any text before or after the JSON.
\end{itemize}

\vspace{0.6em}

Few-shot examples:

\vspace{0.3em}

[FEW-SHOT EXAMPLE 1]

[FEW-SHOT EXAMPLE 2]

\vspace{0.6em}

User input:

\vspace{0.3em}

Discussion:

<conversation>

\vspace{0.4em}

Argument units:

<argument\_units\_json>

\vspace{0.6em}
\rmfamily
\hrule
\end{minipage}
\caption{Prompt template used for relation prediction and relation type classification.}
\label{fig:prompt_relations}
\end{figure}

\section{Quantitative Error Analysis Details}
\label{app:qua-err}

We observe six types of structural errors in the datasets and across configurations. False positive units are predicted units that do not exist in the ground truth structures, likely due to segmentation errors. False negative units are non-predicted units that exist in ground truth structures but were not predicted by the models. These are observed to be more implicit arguments that require human-level reasoning and contextual understanding to be identified.

\begin{figure}[t]
    \centering
    \includegraphics[width=0.55\textwidth]{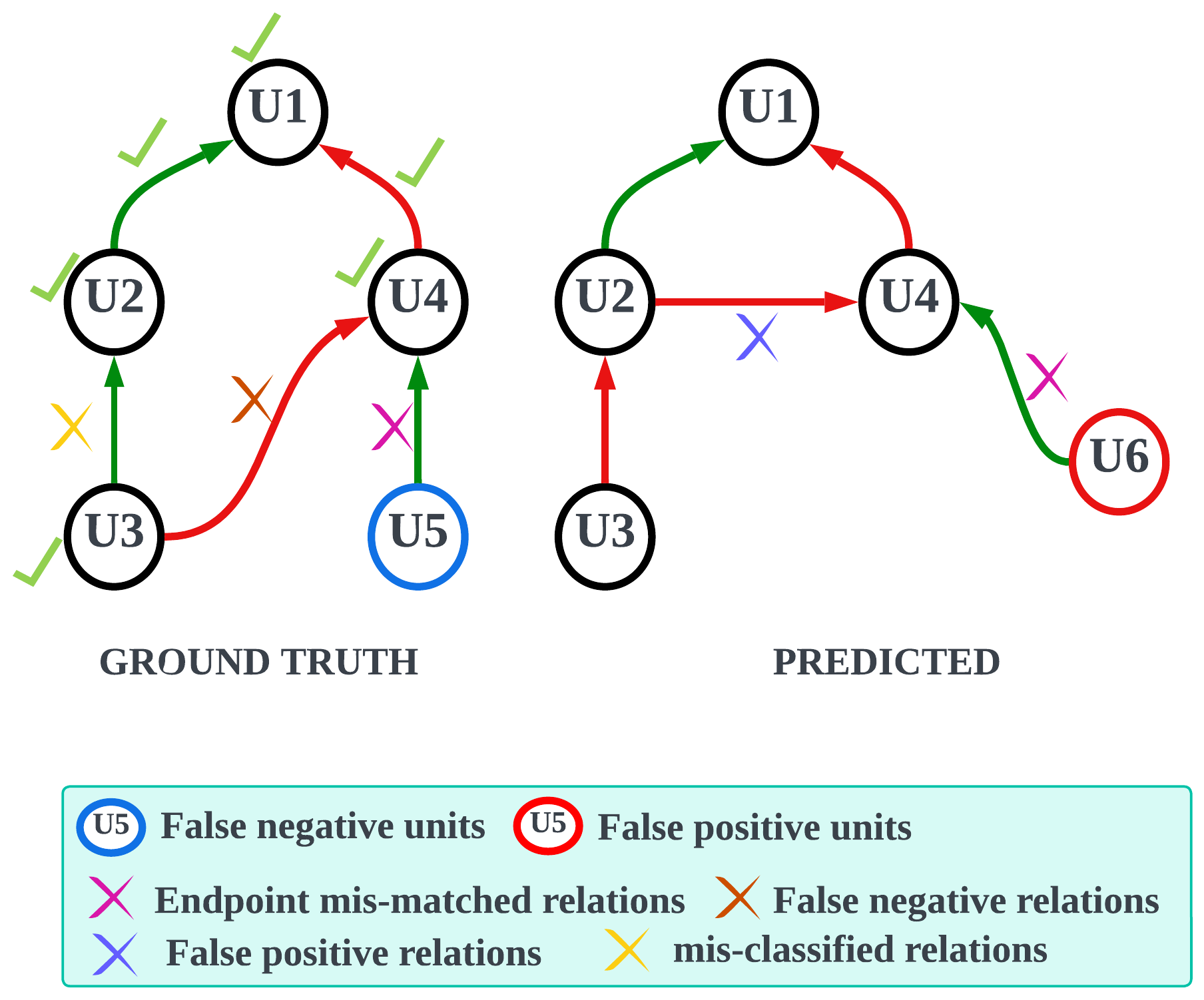}
    \caption{Visualization of the six types of structural errors observed in our error analysis between the ground truth structures and the predicted structures. The green edges represent support relations; red edges represent attack relations; U1-6 are argument units.}
    \label{fig:err_vis}
\end{figure}

In terms of relation errors, we identify four types. Endpoint mis-matched relations are relations that exist either in ground truth or in predicted structure but can not be evaluated because either source or the target unit is false, i.e., all relations that are linked to false positive or false negative units are automatically categorized as endpoint mis-matched relations by us. False positive relations are predicted relations between two true positive units that do not exist in ground truth structure. False negative relations are non-predicted relations between two true positive units that do exist in ground truth structure but missing in predicted structures. Finally, mis-classified relations are relations that have been correctly identified between two true positive units but has been mis-labelled (incorrect support or attack label). Figure~\ref{fig:err_vis} presents a visualization of the six structural errors.

In Table~\ref{tab:error-counts} we list the quantitative values of the six structural errors observed in the structure across the datasets and the configuration. We observe that for QT30 and RIP1 corpora, LLM configurations are generally more liberal, overly predicting units, leading to higher false-positive units and relatively lower false-negative units. But for US2016reddit the reverse behavior is observed, potentially due to this dataset emerging from social media conversations that are more argumentative and more implicit in nature. 

Studying the relation errors, we do have clear indications that all LLMs across the three corpora behave very conservatively with significantly higher false negative relations than false positive relations. Another main failure mode is the endpoint mis-matched relations. Higher false-positive units or false-negative units naturally leads to higher endpoint mis-matched relations. We see the clear dependency between units and the relations where a superior unit alignment strategy allows for reduced endpoint mis-matched relations.

\begin{table}[t]
\centering
\begin{tabular}{lrrrrrr}
\toprule
MC & U$^{+}$ & U$^{-}$ & E & R$^{+}$ & R$^{-}$ & C \\
\midrule
\multicolumn{7}{l}{\textbf{QT30-test}} \\
FT$_{ms}$ & 1163 & 1022 & 1836 & 265 & 934 & 8 \\
Q0$_{ss}$ & \textbf{647} & 1178 & 472 & 71 & 978 & 3 \\
Q3$_{ss}$ & 883 & 630 & 851 & 207 & 825 & 13\\
Q5$_{ms}$ & 853 & 640 & 799 & 228 & 812 & 14\\
Q5$_{ss}$ & 982 & 569 & 921 & 204 & 763 & 18 \\
D5$_{ms}$ & 778 & \textbf{297} & 825 & 168 & 467 & 8 \\
D5$_{ss}$ & 1378 & 490 & 1576 & 246 & 728 & 26 \\

\addlinespace[2pt]
\multicolumn{7}{l}{\textbf{US2016reddit}} \\
FT$_{ms}$ & 907 & 1095 & 2018 & 401 & 1178 & 32 \\
Q0$_{ss}$ & 528 & 1026 & 381 & 131 & 1176 & 19 \\
Q3$_{ss}$ & 592 & 667 & 553 & 267 & 1046 & 29 \\
Q5$_{ms}$ & 476 & 732 & 474 & 312 & 1063 & 23 \\
Q5$_{ss}$ & 588 & 630 & 561 & 250 & 1037 & 26\\
D5$_{ms}$ & \textbf{378} & 285 & 394 & 217 & 537 & 22 \\
D5$_{ss}$ & 779 & 548 & 901 & 364 & 1000 & 51 \\

\addlinespace[2pt]
\multicolumn{7}{l}{\textbf{RIP1}} \\
FT$_{ms}$ & 864 & 420 & 1817 & 263 & 406 & 19 \\
Q0$_{ss}$ & 723 & 569 & 378 & 12 & 448 & 0 \\
Q3$_{ss}$ & 630 & 486 & 447 & 35 & 432 & 4 \\
Q5$_{ms}$ & 476 & 473 & 388 & 51 & 427 & 6 \\
Q5$_{ss}$ & 683 & 437 & 443 & 44 & 420 & 7 \\
D5$_{ms}$ & 745 & 451 & 672 & 71 & 421 & 10 \\
D5$_{ss}$ & 890 & 415 & 927 & 54 & 409 & 6 \\

\bottomrule
\end{tabular}

\caption{
\textbf{Quantitative Error analysis across model configurations} at 50\% overlap threshold using seed 42 for each dataset/best configurations.
U$^{+}$ and U$^{-}$ denote false positive and false negative argument units.
E denotes endpoint-mismatched relations.
R$^{+}$ and R$^{-}$ denote false positive and false negative relations.
C denotes mis-classified relation types.
}
\label{tab:error-counts}
\end{table}



\end{document}